\def\year{2020}\relax
\documentclass[letterpaper]{article} 
\usepackage{aaai20}  
\usepackage{times}  
\usepackage{helvet} 
\usepackage{courier}  
\usepackage[hyphens]{url}  
\usepackage{graphicx} 
\usepackage{graphicx}  
\newcommand{\citet}[1]{\citeauthor{#1} \shortcite{#1}} 
\newcommand{\citep}{\cite}

\usepackage{amsmath}
\usepackage{amsthm}
\usepackage{amssymb}
\usepackage{multirow}
\usepackage{multicol}
\usepackage{booktabs}

\newcommand{\vect}[1]{\mathbf{#1}}
\newcommand{\matr}[1]{\mathbf{#1}}

\newcommand{\vs}[0]{\vect{s}}

\newcommand{\vx}[0]{\vect{x}}
\newcommand{\vy}[0]{\vect{y}}
\newcommand{\vz}[0]{\vect{z}}
\newcommand{\vzero}[0]{\vect{0}}

\newcommand{\vtheta}{\boldsymbol{\theta}}

\newcommand{\loss}{\mathcal{L}}

\newcommand{\E}{\mathbb{E}}

\newcommand{\N}{\mathcal{N}}

\newcommand{\eye}{\matr{I}}

\newcommand\bern{\text{Bern}}

\newcommand\kld{\text{D}_{\textsc{kl}}}

\nocopyright
\title{Text Modeling with Syntax-Aware Variational Autoencoders}
\author{Yijun Xiao, William Yang Wang\\
University of California, Santa Barbara\\
  {\{yijunxiao,william\}@cs.ucsb.edu} 
}
 
\begin{document}

\maketitle

\begin{abstract}
Syntactic information contains structures and rules about how text sentences are arranged. Incorporating syntax into text modeling methods can potentially benefit both representation learning and generation. Variational autoencoders (VAEs) are deep generative models that provide a probabilistic way to describe observations in the latent space. When applied to text data, the latent representations are often unstructured. We propose syntax-aware variational autoencoders (SAVAEs) that dedicate a subspace in the latent dimensions dubbed syntactic latent to represent syntactic structures of sentences. SAVAEs are trained to infer syntactic latent from either text inputs or parsed syntax results as well as reconstruct original text with inferred latent variables. Experiments show that SAVAEs are able to achieve lower reconstruction loss on four different data sets. Furthermore, they are capable of generating examples with modified target syntax.
\end{abstract}

\section{Introduction}

Deep generative models, such as variational autoencoders (VAEs) \cite{kingma2013auto,rezende2015variational} and generative adversarial networks (GANs) \cite{goodfellow2014generative}, have made impressive advances in vision tasks \cite{radford2015unsupervised,gregor2015draw,yan2016attribute2image,mansimov2015generating}. Among the two, VAE introduces a practical training technique for deep neural network generative models with latent variables. They assume the following data generation process: a latent representation is generated from a given distribution, then an output is sampled from a distribution parameterized by a neural decoder conditioned on the latent representation. Variational inference is used to approximate the true posterior distribution with a deep inference network and training of the whole system is enabled through simple stochastic gradient descent (SGD) \cite{robbins1985stochastic}.

Recent attempts on using VAEs for generic text generation \cite{bowman2016generating,yang2017improved,hu2017toward,xiao2018dirichlet,kim2018semi} show promising results but it remains a challenging task as the models are required to capture complex semantic structures underlying sentences. Due to the smoothness of the latent space, one advantage of applying VAE on text generation is that the model is able to generate text examples from continuous samples or even interpolations in the latent space. However, latent representations are often unstructured. It is hard to assign physical meanings to latent space dimensions and their roles during the generation process are often entangled. Semi-supervised VAEs \cite{kingma2014semi}, to some extent, attack this issue by considering categorical attributes of sentences as latent variables. \citet{hu2017toward} also adopt a similar approach and add extra discriminators to encourage generation of text with specified attributes. These methods work well when the targeted attributes are in the forms of simple distributions such as Gaussian or categorical distributions. When the attribute of interest is of discrete structures, these methods are not directly applicable.

Syntactic structures (e.g. constituency parses) are inherent structured attributes of any given text sentences. In the context of natural language understanding (NLU) and generation (NLG), syntax-awareness generally can be either incorporated in encoders to improve representation learning and other downstream tasks \cite{bastings2017graph,li2018unified,strubell2018linguistically}, or added to decoders to restrict generated text to conform to given syntactic structures. In the second case, one approach is to explicitly add structure restrictions during the generation process \cite{dai2018syntax,kusner2017grammar,liu2018treegan,rabinovich2017abstract,yin2017syntactic}. This requires predefined grammar rules for the model to impose correct restrictions. Another approach is to encode syntactic forms into dense representations and feed them to the decoders to bias generation \cite{iyyer2018adversarial}. This approach has the benefit of not relying on predefined grammars but has the drawback of requiring target syntax to perform generation. 

In this paper, we propose a syntax-aware variational autoencoder (SAVAE) that dedicates a subspace in the latent dimensions to represent syntactic structures. The model is able to make inference of this syntactic latent variable from either text inputs or any given surface syntactic forms. The syntactic latent variable is fed into the decoder to guide the generation process. We conduct experiments on four data sets and find our proposed approach improves reconstruction compared to models without syntax-awareness. Further analyses show that our model is also able to infer syntactic structures from input sentences for short documents as well as generate text that conforms to specified syntax.
In summary, our major contributions are:
\begin{itemize}
\item We present a syntax-aware variational autoencoder that utilizes and encodes syntactic information of input text.
\item We empirically show that syntax-awareness helps the proposed model to achieve better reconstruction on four data sets.
\item Our model is able to infer syntactic structures from input sentences for short documents as well as generate text with modified syntax.
\end{itemize}

\section{Related Work}
\label{sec:related}
In this section, we review VAE for text modeling and efforts on syntax-aware text generation.

\subsection{Variational Autoencoder}
Variational autoencoders (VAEs) \cite{kingma2013auto,rezende2015variational} are deep generative models that are composed with encoder and decoder networks. An input example is encoded into a latent representation before getting reconstructed back from the latent space. Since its proposal, VAE has been widely adopted as generative model for images \cite{gregor2015draw,yan2016attribute2image,mansimov2015generating}. There are studies of variational inference on other tasks such as machine translation \cite{zhang2016variational}, knowledge graph reasoning \cite{zhang2018variational,chen2018variational}, topic modeling \cite{srivastava2017autoencoding}, and dialogue systems \cite{serban2017hierarchical,zhao2017learning}.

\citet{bowman2016generating} first explore the use of VAE for text modeling. They uncover the training difficulties when using long short-term memory (LSTM) \cite{hochreiter1997long} decoders to reconstruct text inputs. They find that the training often collapses to ignore the latent variable and reconstruct text sequences using only a language model. They propose to use annealing scheme and word dropping to ease the issue. \citet{miao2016neural} apply variational autoencoder on bag-of-word representations of documents. Improvements are made but the model is not able to generate text sequences. \citet{yang2017improved} argue that the training difficulties are mainly due to the use of strong LSTM decoders. They adopt dilated convolutional neural networks \cite{yu2016multi} as decoders and observe improvements over LSTM decoders. \citet{hu2017toward} couple VAE with discriminators to encourage the generated text to have certain specified attributes. They experiment with sentiment and tense attributes and show their model is able to perform controlled text generation. \citet{xiao2018dirichlet} use a Dirichlet latent variable to represent document topic distributions and show it enhances reconstruction and representation learning abilities of VAE. This paper focuses on utilizing syntactic information in a VAE model to help improve text modeling and generation.

\subsection{Syntax-Aware Text Generation}
Syntax-awareness is beneficial for many NLU tasks \cite{strubell2018linguistically,li2018unified}. For text generation, there are also various efforts on restricting output syntax. One approach is to incorporate structure restrictions explicitly into the generative model \cite{kusner2017grammar,dai2018syntax,liu2018treegan}. For example, \citet{kusner2017grammar} propose a grammar variational autoencoder in which context-free grammars are incorporated into the decoder to generate valid molecule structures. This line of work makes use of grammar rules to enforce syntax validity during generation. \citet{iyyer2018adversarial} experiment syntax control in the context of paraphrase generation. Target syntactic form is encoded and fed into the paraphrase generator to bias the generation. Grammar rules are no longer required but one drawback of this method is that target syntactic structures have to be provided during training and testing. There is no way to infer most probable structures. In this study, we propose a VAE model that is not only able to generate text with guidance of target syntax, but also capable of making inference of possible syntactic structures if none is provided.

Another closely related research field is controllable text generation. The efforts on separating text contents from other attributes such as sentiment and writing style have surged drastically since the introduction of sequence-to-sequence architectures used for natural language generation (NLG).
Particularly, the research on learning style transfer from non-parallel text data is gaining attention \cite{mueller2017sequence,shen2017style,hu2017toward,fu2018style,yang2018unsupervised,zhao2018adversarially}. 
Our work shares a similar spirit but focuses on imposing discrete sequence structures to the generated text.

\section{Methods}
\label{sec:method}
We introduce syntax-aware variational autoencoders (SAVAEs) for text modeling. The proposed model aims to improve text VAE by incorporating syntactic information at training time. Ideally, the model not only could utilize the provided syntax in training, but also has the capability to generate new examples with or without syntax supervision at inference time. In other words, when receiving a new sentence input, the VAE model is capable of: 1. reconstructing the sentence with or without original syntactic form; 2. inferring syntax from the input text; 3. controlling the structure of the generated text by feeding target syntactic form.

\begin{figure}[t]
\centering
\begin{minipage}[b]{0.14\textwidth}
\centering
\includegraphics[width=\textwidth]{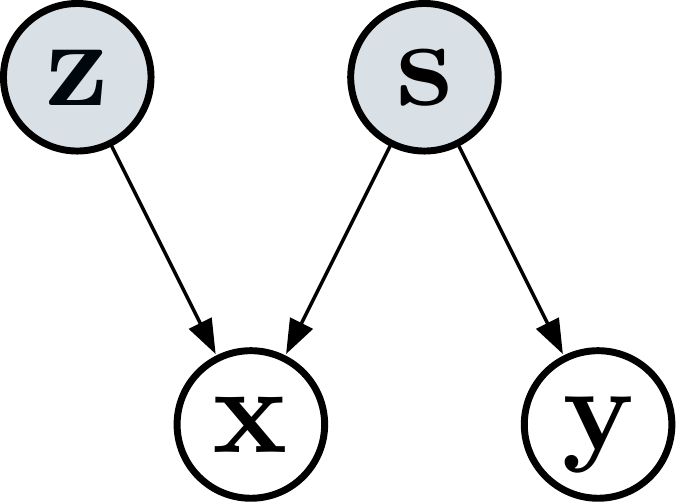}
\\
(a)
\end{minipage}
\hfill
\begin{minipage}[b]{0.14\textwidth}
\centering
\includegraphics[width=\textwidth]{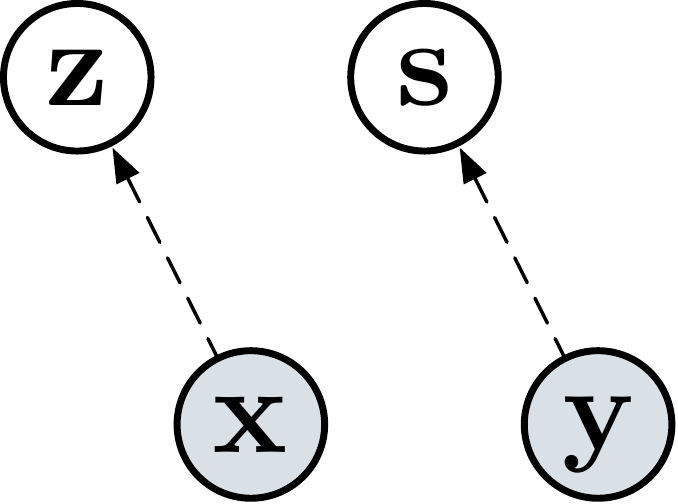}
\\
(b)
\end{minipage}
\hfill
\begin{minipage}[b]{0.14\textwidth}
\centering
\includegraphics[width=\textwidth]{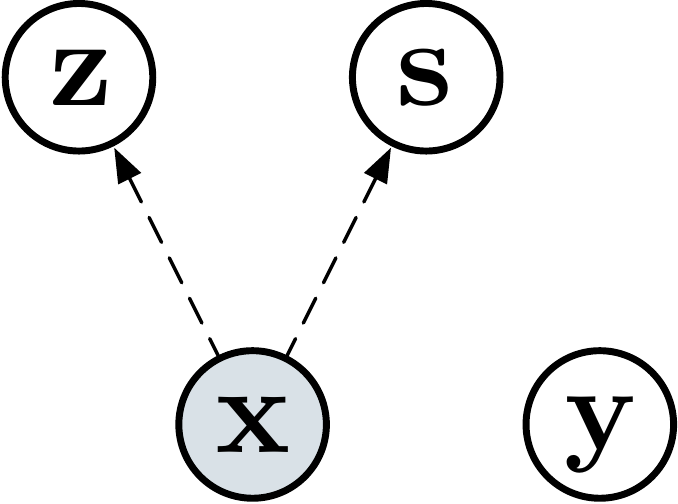}
\\
(c)
\end{minipage}
\caption{Graphical illustration of (a) the generative model (b) the recognition model when both $\vx$ and $\vy$ are observed and (c) the recognition model when only $\vx$ is observed. $\vx$ is text tokens, $\vy$ is its syntax; $\vz,\vs$ are content/syntactic latent variables respectively. Solid lines are the shared generative model; dashed lines are the respective recognition models; shaded nodes represent observed variables.}
    \label{fig:models}
\end{figure}

\subsection{Model Overview}
We build our model based on the variational inference framework which has been used for text modeling \cite{bowman2016generating}. The vanilla VAE generates sentence $\hat{\vx}$ conditioned on the latent representation $\vz$. Conditional generation in the VAE context normally takes the form of either conditional VAE (CVAE) \cite{sohn2015learning} or semi-supervised VAE \cite{kingma2014semi}. CVAE requires the condition to be present both in training and testing. Semi-supervised VAE, on the other hand, models the conditions / attributes as latent variables and performs marginalization whenever a variable is not observed during training. However, due to the discrete nature of syntax, we are not able to model it using any of the simple distributions. Therefore the semi-supervised VAE framework described in \cite{kingma2014semi} is not directly applicable. 

To model the syntactic information stored in the latent representation, we use an additional latent variable $\vs$ by making the assumption that syntax surface form $\vy$ is generated from this particular variable. A simple example of $\vy$ can be the part-of-speech (POS) tags of the input text. To effectively learn $\vs$, we need the surface form syntax $\vy$ during training. There has to be enough supervision in the training to successfully differentiate $\vs$ from $\vz$. The generation of a sentence $\hat{\vx}$ is conditioned on both $\vz$ and $\vs$. Ideally, the syntax of $\hat{\vx}$ should match the information stored in $\vs$.

One issue associated with the model is the choice of recognition model for syntactic latent variable $\vs$. If we choose to infer $\vs$ based on both the sentence $\vx$ and the syntax $\vy$, we are constrained to provide both whenever inference is required. In this paper, we employ two recognition models simultaneously. $\vs$ can be inferred from either $\vx$ or $\vy$ independently. This modeling choice not only enables us to infer $\vs$ when syntax is not available but also makes control with altered syntax possible. Figure \ref{fig:models} illustrates the generative model and two recognition models used in our approach. 

\begin{figure}[t]
\centering
\includegraphics[width=0.45\textwidth]{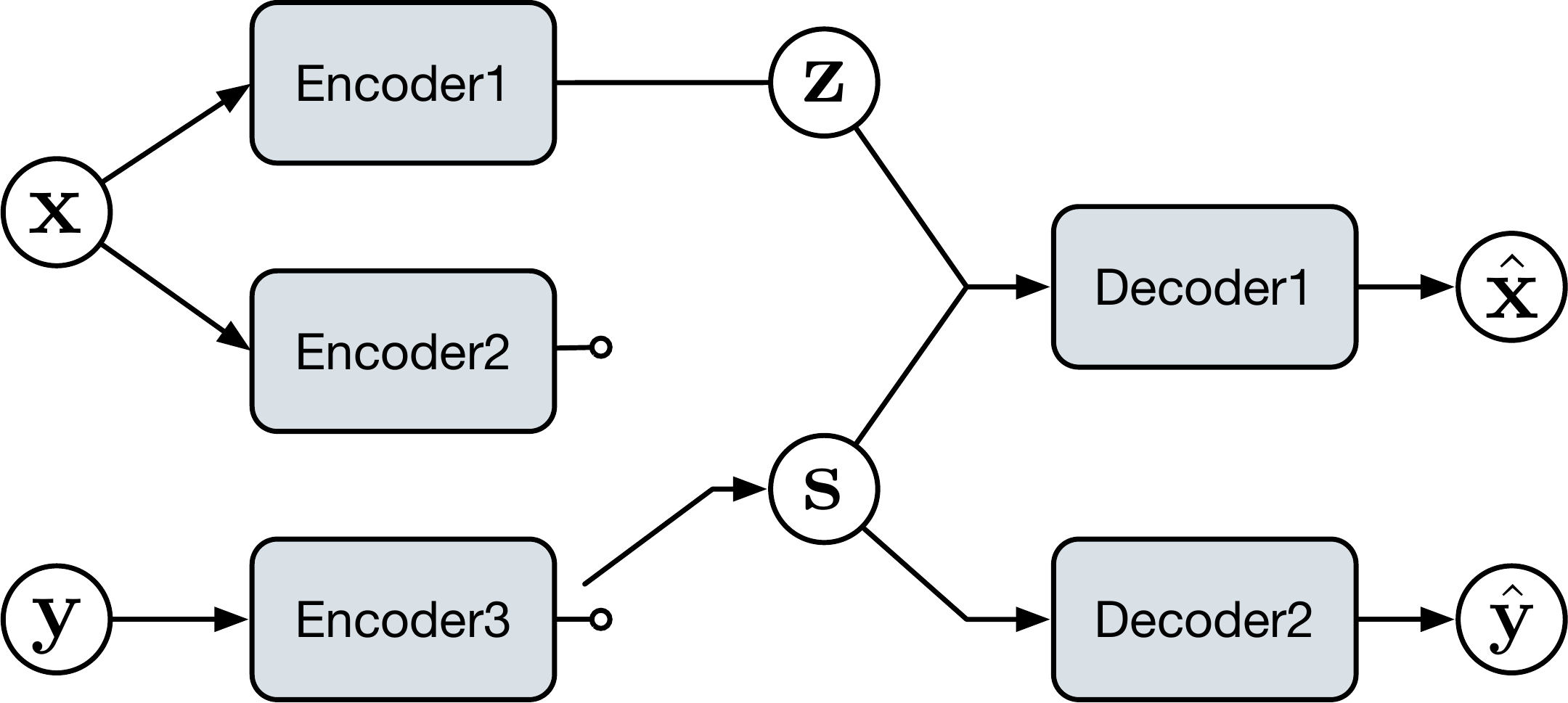}
\caption{Network structure for the model. Empty circles denote switch nodes. At training time, the switch is randomly selected. At inference time, the switch is manually set depending on the use cases.}\label{fig:model_structure}
\end{figure}

The overall model structure is shown in Figure \ref{fig:model_structure}. There are three encoders and two decoders: two of the encoders encode sentence $\vx$ into latent representations $\vz$ and $\vs$ respectively; one encodes syntax $\vy$ into $\vs$; two decoders are used to generate sentence $\hat{\vx}$ and syntax surface form $\hat{\vy}$ respectively. Notice the switch in the diagram. During training, the model randomly selects outputs from \texttt{Encoder2} and \texttt{Encoder3} to make inference of $\vs$. A semi-supervised learning setting is also possible due to the use of two encoders. It is worth stressing that enough supervision from the original syntax $\vy$ through \texttt{Decoder2} is necessary in learning a suitable syntactic latent $\vs$. 

At test time, whenever $\vy$ is not available for a specific input sentence $\vx$, \texttt{Encoder2} is used to infer $\vs$; we could also choose to use an altered syntax and obtain latent $\vs$ by feeding the altered syntax through \texttt{Encoder3}.

\subsection{Model Structure}
In this section, we describe our model in detail, by presenting the learning mechanisms.

There are four variables involved in the model. Text token sequence $\vx=\{x_1,x_2,\cdots,x_T\}$, syntax token sequence $\vy=\{y_1,y_2,\cdots,y_K\}$, general latent variable $\vz$ and syntactic latent variable $\vs$. $\vz$ and $\vs$ are sampled from multivariate Gaussian distributions with their priors being the standard Normal distributions:
\begin{align}
    p(\vz)&={}\N(\vzero_{d_z},\eye_{d_z}) \nonumber\\
    p(\vs)&={}\N(\vzero_{d_s},\eye_{d_s})
\end{align}
where $d_z$ and $d_s$ denote the dimensions of $\vz$ and $\vs$ respectively. A generator $G_x$ is used to generate the text tokens $\vx$ from the combined latent code $(\vz,\vs)$:
\begin{align}
    \vx \sim G_x(\vz, \vs)&={}p_{G_x}(\vx|\vz,\vs) \nonumber \\
    &={}\prod_{t=1}^Tp(x_t|\vx_{<t},\vz,\vs)
\end{align}
where $\vx_{<t}$ denotes the collection of tokens before step $t$. Similarly, the generative story for syntax surface form involves a decoder $G_y$:
\begin{align}
    \vy \sim G_y(\vs)&={}p_{G_y}(\vy|\vs) \nonumber \\
    &={}\prod_{k=1}^Kp(y_k|\vy_{<k},\vs)
\end{align}

Given an input sentence $\vx$ and syntax $\vy$, probabilistic encoders $E_z$ and $E_s$ are used to make inference of the latent representations $\vz$ and $\vs$:
\begin{align}\label{equ:posterior_zs}
    \vz \sim{} & E_z(\vx)={} q_{E_z}(\vz|\vx) \nonumber\\
    \vs \sim{} & E_s(\vx,\vy) \nonumber\\
    &={} \alpha\cdot q_{E_s}(\vs|\vx)+(1-\alpha)\cdot q_{E_s}(\vs|\vy)
\end{align}
where $\alpha$ controls the weights of inferring $\vs$ from $\vx$ and $\vy$. 
When both $\vx$ and $\vy$ are observed, the log-likelihood of the joint distribution of $\vx$ and $\vy$ can be written as:
\begin{align}\label{equ:elbo_xy}
    \log p(\vx,\vy) ={} & \log \int_{\vz,\vs} p(\vx,\vy,\vz,\vs)d\vz d\vs \nonumber\\
    \ge{} & \E_{q(\vz,\vs|\vx,\vy)}\left[\log p_{G_x}(\vx|\vz,\vs)\right] \nonumber\\
    & +\E_{q(\vs|\vx,\vy)}\left[p_{G_y}(\vy|\vs)\right]\nonumber\\
    & -\kld\left(q_{E_z}(\vz|\vx)\|p(\vz)\right) \nonumber\\
    & -\kld\left(q(\vs|\vx,\vy)\|p(\vs)\right)
\end{align}
where $\kld(\cdot\|\cdot)$ denotes the Kullback-Leibler (KL) divergence between two distributions. Here, we use mean field approximation to the true posterior distribution $p(\vz,\vs|\vx,\vy)$ by decomposing the variational distribution as:
\begin{align}
    q(\vz,\vs|\vx,\vy)={}q_{E_z}(\vz|\vx)\cdot q(\vs|\vx,\vy)
\end{align}

Similarly, in the case where only $\vx$ is observed, the log-likelihood of the marginal distribution is:
\begin{align}\label{equ:elbo_x}
    \log p(\vx)={} & \log \int_{\vy,\vz,\vs} p(\vx,\vy,\vz,\vs)d\vy d\vz d\vs \nonumber\\
    ={}& \log \int_{\vz,\vs} p(\vx,\vz,\vs)d\vz d\vs \nonumber\\
    \ge{}& \E_{q(\vz,\vs|\vx)}\left[p_{G_x}(\vx|\vz,\vs)\right] \nonumber\\
    &-\kld\left(q_{E_z}(\vz|\vx)\|p(\vz)\right) \nonumber\\
    &-\kld\left(q_{E_s}(\vs|\vx)\|p(\vs)\right)
\end{align}
Again, mean field approximation is applied on the posterior distribution $p(\vz,\vs|\vx)$.

Lower bounds of the data log-likelihood derived in Equation \ref{equ:elbo_xy},\ref{equ:elbo_x} are often referred to as evidence lower bounds (ELBOs). It is obvious that ELBO derived in Equation \ref{equ:elbo_x} is exactly the same with the vanilla VAE ELBO. Therefore, the proposed model collapses to a vanilla VAE when no syntax supervision is provided at training time. 

Let $\vtheta_E$ and $\vtheta_G$ denote the parameters of the encoders and decoders respectively. Learning aims to minimize the negative ELBO. Intuitively, it minimizes the reconstruction errors of both $\vx$ and $\vy$ at the same time regularizes the variational distributions to be close to the priors.
\begin{align}\label{equ:loss}
    \loss(\vtheta_E,\vtheta_G;\vx,\vy)={}&\kld\left(q_{E_z}(\vz|\vx)\|p(\vz)\right) \nonumber\\
    &+\kld\left(q(\vs|\vx,\vy)\|p(\vs)\right)\nonumber\\
    &-\E_{q(\vz,\vs|\vx,\vy)}\left[\log p_{G_x}(\vx|\vz,\vs)\right] \nonumber\\
    &-\E_{q(\vs|\vx,\vy)}\left[p_{G_y}(\vy|\vs)\right]
\end{align}

To draw samples from $q(\vs|\vx,\vy)$, we follow the following procedure:
\begin{align}
    a&\sim\bern(\alpha) \nonumber\\
    \vs&\sim a\cdot q_{E_s}(\vs|\vx)+(1-a)\cdot q_{E_s}(\vs|\vy)
\end{align}
where $\bern(\cdot)$ is the Bernoulli distribution. KL divergence from the posterior to the prior for latent variable $\vs$ can be estimated using Monte Carlo estimation as:
\begin{multline}
    \kld\left(q(\vs|\vx,\vy)\|p(\vs)\right) \approx{} \\
    \frac{1}{K}\sum_{k=1}^K \log{q(\vs_k|\vx,\vy)}-\log{p(\vs_k)}
\end{multline}
with $\vs_k$ drawn from $q(\vs|\vx,\vy)$.

The model can be trained end-to-end by back-propagating the loss defined in Equation \ref{equ:loss} through all model components. Semi-supervised learning objective can also be derived in a similar manner. In this paper, as the labeling of syntax $\vy$ is automatic, we apply fully supervised setting in the experiments.

\section{Experiments}
\label{sec:experiments}
We conduct experiments on four different data sets with various input lengths. We use part-of-speech tags of the input sequences as the added syntactic information. We evaluate the model capabilities of reconstructing sentences, inferring syntax from text, and manipulating syntax during generation. In this section, we describe the data sets, detailed setups, and experimental results.
\subsection{Data Sets}
We choose four different data sets in our experiments. The Penn Treebank (PTB) \cite{marcus1993building} and the BookCorpus (BC) \cite{zhu2015aligning} have relatively short sentences, while Yahoo Answer and Yelp15 review consist of relatively long documents. 
For BC data set, we randomly sample 200k sentences for training, 10k for validation and 10k for test. Subsets of Yahoo and Yelp data used in \cite{yang2017improved} are adopted. Each subset contains 100k documents for training, 10k for validation and 10k for test. Class labels in Yahoo and Yelp data are not used in the experiments. Statistics of the data sets are summarized in Table \ref{tab:data}.

\begin{table}[t]
    \centering
    \begin{tabular}{crrr}
    \toprule
         \bf Data & \bf Size & \bf Average \#w & \bf Vocabulary\\
         \midrule
         PTB & 42k & 21 & 10k \\
         BC & 200k & 8 & 43k \\
         Yahoo & 100k & 78 & 20k \\
         Yelp & 100k & 96 & 20k \\
         \bottomrule
    \end{tabular}
    \caption{Key statistics of the training data sets. \#w represents document length by number of tokens.}
    \label{tab:data}
\end{table}

To obtain syntactic information, we use spaCy v2.0\footnote{https://spacy.io/} to parse all data examples and obtain part-of-speech (POS) tags. The POS sequences are considered as syntax inputs in all experiments.

\begin{table*}[t!]
\centering
\begin{tabular}{llllllllll}
\toprule
 \multirow{2}{*}{\bf Model} & \multicolumn{2}{c}{\bf PTB} & \multicolumn{2}{c}{\bf BC} & \multicolumn{2}{c}{\bf Yahoo} & \multicolumn{2}{c}{\bf Yelp} \\
 & NLL & PPL & NLL & PPL & NLL & PPL & NLL & PPL \\
  \midrule
  LSTM-LM \shortcite{mikolov2011extensions} & 101.8 & 104.2 & 37.4 & 64.0 & 334.9 & 66.2 & 362.7 & 42.6\\ 
    LSTM-VAE \shortcite{bowman2016generating} &  102.0 & 104.8 & 28.0 & 22.5 & 337.4 & 68.3 & 372.2 & 47.0 \\
CNN-VAE \shortcite{yang2017improved}& - & - & - & - & 333.9 & 65.4 & \textbf{361.9} & \textbf{42.3}\\
\midrule
SA Baseline 1 & 95.3 & 77.4 & 26.1 & 18.1 & \textbf{325.3} & \textbf{58.7} & 362.8 & 42.7 \\
SA Baseline 2 & 94.8 & 75.5 & 23.6 & 13.7 & 331.9 & 63.8 & 364.3 & 43.4  \\
LSTM-SAVAE (Ours) & \textbf{92.9} & \textbf{69.5} & \textbf{23.2} & \textbf{13.2} & 331.2 & 63.2 & 363.8 & 43.1\\

\bottomrule
\end{tabular}
\caption{Language modeling results on the test sets. Lower is better. For SAVAE, syntax is only provided at training while for SA baseline syntactic information is presented in both training and evaluation. SAVAE has better reconstruction qualities across all four data sets compared to its non-syntax-aware counterpart.} 

\label{tab:lm}
\end{table*}

\subsection{Experiment Settings}
Input vocabulary is capped at 20k. Input embedding size is 200 for text tokens and 50 for syntax tokens. Single-layer LSTM models are employed in all encoders and decoders. Hidden sizes for \texttt{Encoder1}, \texttt{Encoder2}, and \texttt{Decoder1} (refer to Figure \ref{fig:model_structure}) are set to 512; \texttt{Encoder3} and \texttt{Decoder2} have a hidden size of 128. Dimensions of latent variables $\vz$ and $\vs$ are 32. Dropout ratios for all encoders and decoders are $0.2$.

\texttt{Decoder1} state is initialized with $\vz$; syntactic latent $\vs$ is concatenated to inputs at each decoding step. We also experiment initialization and step feeding with both latent variables, but the results are similar. \texttt{Decoder2} is initialized with $\vs$ and no step feeding is applied. The probabilities of inferring $\vs$ from \texttt{Encoder2} and \texttt{Encoder3} during training are set to $0.5$ each.

Models are trained with Adam \cite{kingma2014adam} with learning rate [1e-3, 3e-3]. $\beta_1,\beta_2,\epsilon$ are kept to default values specified in the original paper. No further parameter tuning is conducted. We use batch size of 64 for PTB and BC, 32 for Yahoo and Yelp data sets. Maximum training epoch is 24 across all experiments. We apply a linear scheduling to anneal the KL weight from 0 to 1.

\subsection{Language Modeling Results}
Language modeling results are shown in Table \ref{tab:lm}. As different ELBOs are used in training, We only report the reconstruction negative log-likelihood (NLL) and the perplexity (PPL) on the test data. A lower NLL and PPL indicates the model is better at reconstructing the input sentences which also shows superior generation quality. 

We compare with LSTM language model (LSTM-LM) \cite{mikolov2011extensions}, LSTM-VAE \cite{bowman2016generating}, and dilated convolutional VAE (CNN-VAE) \cite{yang2017improved}. For SAVAE, original syntactic information is not fed into the model during evaluation. Syntactic latent $\vs$ is inferred from input text $\vx$ only, therefore the comparison is fair. 
We also consider two simple syntax-aware baselines:
\begin{itemize}
    \item SA Baseline 1: syntax token embeddings are concatenated with text token embeddings and a single LSTM encoder is used to encode them into $\vz$. For fair comparison, we double the latent code size for this baseline.
    \item SA Baseline 2: syntactic latent variable $\vs$ is always inferred from $\vx$ during training. This is equivalent to setting $\alpha=1$ in Equation \ref{equ:posterior_zs}.
\end{itemize}  

Theoretically, as syntactic information has to be present in both training and evaluation, SA baseline 1 should outperform other models.

From Table \ref{tab:lm} we observe improvements of the proposed LSTM-SAVAE on all four data sets over the baseline LSTM-VAE model. For Yelp data, CNN-VAE has a lower perplexity by employing a different decoder structure. Our proposed approach is agnostic to decoder structures and can be applied on top of CNN-VAE as well. Interestingly, LSTM-SAVAE outperforms syntax-aware baseline 1 on PTB and BC data sets possibly due to a better generalization for shorter documents.

SA baseline 2 have similar or slightly worse performances across the board compared to LSTM-SAVAE. This indicates that syntactic latent variable $\vs$ can be inferred from $\vx$, but adding the switch makes the framework more flexible and also slightly improves text construction.

It is worth noting that our model performs better on data sets with shorter document lengths. There are two reasons why this might happen. Firstly for longer sequences, the syntactic structure is harder to capture and store in the latent code $\vs$. Because of this, during reconstruction there is less accurate syntactic information provided to the decoder. Secondly, as we adopt an external parser to extract syntactic information from input texts, parsing accuracy for longer sentences is expected to be lower. Therefore the syntactic information used in training is of lower quality which in turn affects the learning of the syntactic latent $\vs$.

\begin{table}[t]
    \centering
    \begin{tabular}{lcccc}
    \toprule
         \multirow{2}{*}{\bf Input} & \multicolumn{3}{c}{\bf Recall @K} & \multirow{2}{*}{\bf Lev} \\
         & Top 1 & Top 3 & Top 10 & \\
         \midrule
         \textsc{(vae)} \\
         syntax $\vy$ & 0.24 & 0.33 & 0.41 & 3.22\\
         \midrule
         \textsc{(savae)} \\
         text $\vx$ & 0.35 & 0.49 & 0.59 & 1.83\\
         syntax $\vy$ & 0.47 & 0.60 & 0.69 & 1.53\\
         \bottomrule
         
    \end{tabular}
    \caption{Evaluation results for generated syntax with different models and inputs on BC data set. Recall @K is the exact match recall rate from top K predictions. Lev is the average Levenshtein distance between gold and inferred syntax tokens.}
    \label{tab:syntax_acc}
\end{table}

\begin{table*}[t!]
\centering
\begin{tabular}{cll}
\toprule
\textsc{input} & \bf Now I must return to the lake. & \bf Everything seemed to happen at once.\\
\textsc{mean} & Now I can get to the door. & Everyone started to get out of here.\\ 
\midrule
\multirow{3}{*}{\textsc{fix} $\vs$} & \it Then I can get to the door. & \it Nobody moved to get out of here.\\
 & \it Maybe he could go to the ground. & \it Someone had to be in there.\\
 & \it Now I can get to the door. & \it We've got to go back home.\\ 
\midrule
\multirow{3}{*}{\textsc{fix} $\vz$} & \it Now, I can do that. & \it Nobody else would have to go.\\
 & \it Now I can help you in the morning. & \it Something had to come in here.\\
 & \it Can I help you in the house? & \it Time to go back to bed.\\
\bottomrule
\end{tabular}
\caption{Examples of generated sentences from the posterior distributions of $\vz$ and $\vs$. $\vs$ is fixed to the mean when sampling from $\vz$ and vice versa. We can observe that when $\vs$ is fixed, generated sentences have exactly the same POS tags.}
\label{tab:rec_samples}
\end{table*}

\begin{table}[h]
    \centering
    \begin{tabular}{cc}
    \toprule
         \bf Setting & \bf Uniq. Syntax \\
         \midrule
         \textsc{vae std.} & 9.4 / 10 \\
         \textsc{savae std.} & 8.6 / 10 \\
         \textsc{savae fix $\vz$} & 8.2 / 10 \\
         \textsc{savae fix $\vs$} & 5.1 / 10 \\
         \bottomrule
    \end{tabular}
    \caption{Average number of unique syntactic structures out of 10 reconstructed text samples on BC test set. Generated samples have lower diversity in syntactic structures when latent $\vs$ is fixed during generation.}
    \label{tab:syntax_var}
\end{table}

\subsection{Quality of Inferred Syntax}
With \texttt{Decoder2} in Figure \ref{fig:model_structure}, the model is capable of reconstructing syntax from latent variable $\vs$. In SAVAE, syntactic latent $\vs$ can be inferred using either the original text $\vx$ or syntax $\vy$. We also train a VAE model on syntax $\vy$ and use the reconstruction results as a baseline.

As decoding steps do not have direct access to the input tokens, there is no way to enforce the generation length. Therefore, we adopt two evaluation metrics: recall rate at top K and Levenshtein distance instead of position-wise accuracy.

\textbf{Recall @K} is the possibility that the gold syntax is within top K predictions. During generation, we fix the sampling of syntactic latent code $\vs$ to its mean value and reconstruct using beam search. Beam size is set to 10 and we keep the top 10 predictions at the end. 

\textbf{Levenshtein distance} (edit distance) measures the number of insertions, deletions, and substitutions needed to transform a string to another. We obtain the best prediction from beam search and compare to the gold syntax. The algorithm used is proposed by \cite{hyyro2001explaining}. Lower is better.

Evaluations are conducted on the BC test data due to its shorter document lengths compared to other three data sets. The results are listed in Table \ref{tab:syntax_acc}. It is interesting to see that training a VAE on $\vy$ itself does not provide a better reconstruction of $\vy$. With the proposed model, the reconstructed syntax is more accurate when inferring syntactic latent $\vs$ from $\vy$. When inferring $\vs$ from text input $\vx$, the accuracy of the predicted syntax is lower but still reasonably good.

\section{Model Analysis}
\label{sec:analysis}
In this section, we evaluate model behaviors by analyzing generated samples in different settings.

\subsection{Reconstruction}
The model is able to reconstruct texts by firstly encoding inputs into latent space and decoding based on the samples from the posterior distributions of $\vz$ and $\vs$. Three different settings are considered: (1) \textsc{mean}: use mean values of the posteriors to do reconstruction; (2) \textsc{fix $\vs$}: fix $\vs$ to the mean and sample $\vz$ from its posterior; (3) \textsc{fix $\vz$}: fix $\vz$ to the mean and sample $\vs$ from its posterior.

Table \ref{tab:rec_samples} shows two examples of reconstruction results. We can observe from the examples that when fixing $\vs$ to the mean, all reconstructed sentences have exactly the same or very similar sentence structures. When reconstructing from samples of $\vs$, there are more variations in syntax.

Further analysis is done to quantify the diversity of generated text sentences with respect to their syntactic structures. In addition to the settings mentioned above, the standard setting (\textsc{std.}) where both latent codes are sampled is also considered. We use spaCy to parse the reconstructed sentences and compute the average number of diverse syntactic structures out of ten samples. Standard setting with VAE is also added for comparison. 

Average numbers of unique syntactic structures in all four settings are listed in Table \ref{tab:syntax_var}. In general, SAVAE generates sentences with lower syntax diversity. Specifically, syntactic latent variable $\vs$ has a strong control over the structures of the generated sentences. When fixing $\vs$, there are on average only 5.1 different syntactic structures out of 10 samples, compared to 8.6 in the standard setting and 8.2 when fixing $\vz$ and sample $\vs$.

\begin{figure}[t]
    \centering
    \includegraphics[width=0.45\textwidth]{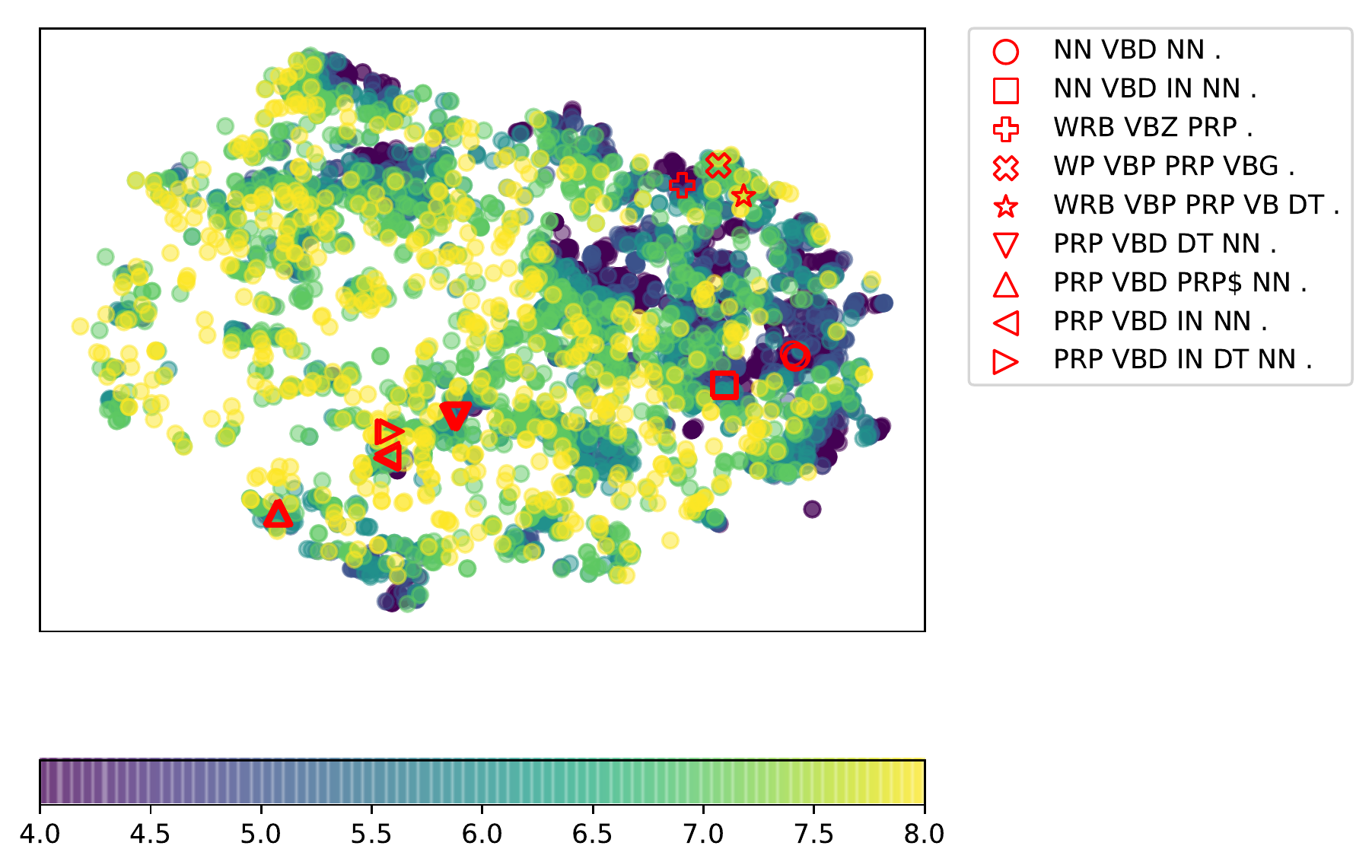}
    \caption{t-SNE scatter plot of the space of $\vs$ inferred from input text $\vx$ on BC dataset. Color represents sentence length. We can observe that sentences with similar POS tags obtain similar latent syntax $\vs$. All test sentences with the given POS sequences are shown in the figure. Best viewed with color.}
    \label{fig:s_space}
\end{figure}

\subsection{Syntactic Latent Space}
To better understand the syntactic latent space, we visualize the syntactic latent $\vs$ on the BC test set using t-SNE \cite{maaten2008visualizing}. $\vs$ is inferred from input text $\vx$. Figure \ref{fig:s_space} shows a scatter plot of t-SNE embeddings of the latent $\vs$. Darker colors indicate shorter input sentences. Embeddings with certain ground-truth POS tag sequences are labeled in red. We can see that for input texts with exactly the same POS tags, the model learns to map them into almost identical points in the syntactic latent space. For texts with similar POS sequences, they are close together in the syntactic latent space.

\subsection{Syntax Modification}
A key advantage of the proposed model is that syntactic latent can be inferred from either input text or syntactic information. This makes modifications of sentence structures possible. We can control the structures of generated sentences by simply feeding the preferred syntactic form $\vy$.

We conduct a probing experiment to quantify this particular capability of the proposed model. We select all test examples in the BC data set where exactly one verb type is detected with the parser. There are 5,776 examples satisfying this requirement. While keeping text inputs the same, we construct five different sets of syntax inputs by replacing all verb types tagged by the parser with [VBD, VBZ, VBP, VBG, VBN]. Syntax modification is accomplished by inferring latent code $\vs$ from the modified syntax inputs and generating sentences conditioned on it.

\begin{figure}[t]
    \centering
    \includegraphics[width=0.42\textwidth]{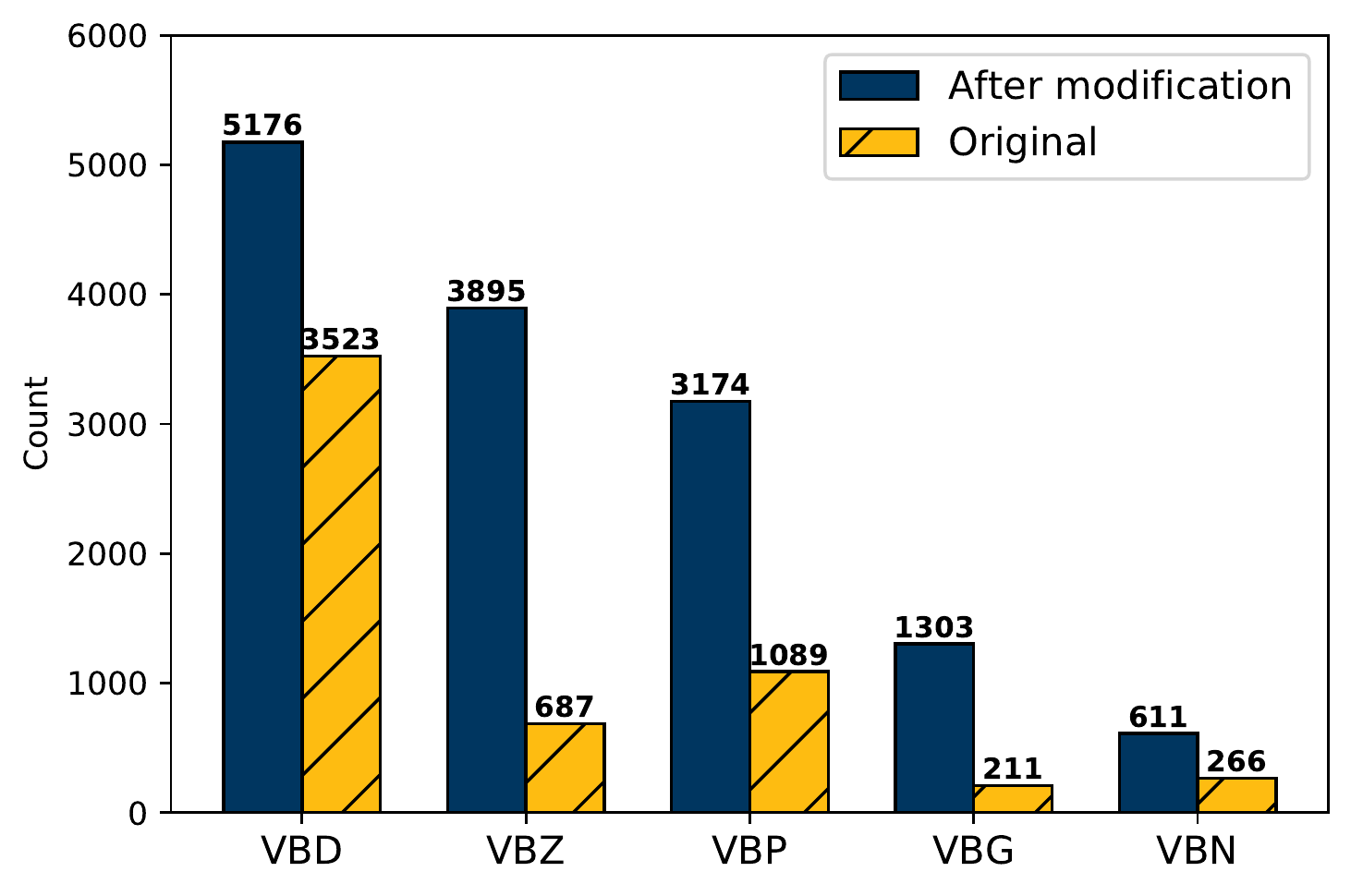}
    \caption{Number of generated examples with each verb type after verb modification compared to the number of examples with each verb type in the original test subset. Total number of examples is 5,776.}
    \label{fig:verb_mod}
\end{figure}

Figure \ref{fig:verb_mod} shows the number of examples with each verb type in the original non-modified test subset compared to the reconstructed set with modified verb types. In the ideal case where total control is achieved, all generated examples should have the specified verb type. Although reconstructions are not perfectly modified for every verb types, the model shows its capability of generating text samples according to specified syntax. 

Table \ref{tab:verb_mod} shows one example of reconstruction with five different modifications. We can see that modifications with VBD, VBG, VBZ, and VBP are (at least partially) successful. It is also interesting to see how the model is trying to generate a VBN in the third example ``His mother was unk'' but fails to find a proper verb.

\begin{table}[t]
    \centering
    \begin{tabular}{cl}
    \toprule
         \multirow{2}{*}{\textsc{input}} &  My heart \texttt{pounded}, my lungs \\
         & \texttt{screamed} like bellows. \\
         \midrule
         \multirow{2}{*}{\textsc{vbd}} & His heart \texttt{skipped}, his voice \\
         & \texttt{laced} with sulfur. \\
         \multirow{2}{*}{\textsc{vbg}} & My name is unk, his eyes \\ 
         & \texttt{flaring} with uncertainty. \\
         \multirow{2}{*}{\textsc{vbn}} & His mother was unk, her voice \\
         & laced with sulfur. \\
         \multirow{2}{*}{\textsc{vbz}} & My name \texttt{is} Sheba, my specialty \\
         & \texttt{is} unk.\\
         \multirow{2}{*}{\textsc{vbp}} & My name is Sheba, its going to \\
         & \texttt{be} exact. \\
         \bottomrule
    \end{tabular}
    \caption{Example results of modifying verb types. Modifications are successful with VBD, VBG, VBZ, and VBP while unsuccessful with VBN.}
    \label{tab:verb_mod}
\end{table}
\section{Conclusions}
\label{sec:conclusions}

In this study, we propose a VAE-based deep generative model for text modeling that systematically incorporates syntactic structures within the variational inference framework. The model learns a meaningful latent subspace to store syntactic information. The learned syntactic latent variable is used to guide generation of sentences towards the encoded syntactic form.
Empirical experiments show that syntax-awareness helps the model achieve lower text reconstruction errors on four data sets. Furthermore, the model is able to reliably predict syntactic structures from short sentences. It is also capable of generating sentences that conform to modified syntactic structures. The same framework can be applied to model objects with any structured attributes.

\bibliographystyle{aaai}
\bibliography{aaai20}

\begin{thebibliography}{}

\bibitem[\protect\citeauthoryear{Bastings \bgroup et al\mbox.\egroup
  }{2017}]{bastings2017graph}
Bastings, J.; Titov, I.; Aziz, W.; Marcheggiani, D.; and Simaan, K.
\newblock 2017.
\newblock Graph convolutional encoders for syntax-aware neural machine
  translation.
\newblock In {\em Proceedings of the 2017 Conference on Empirical Methods in
  Natural Language Processing},  1957--1967.

\bibitem[\protect\citeauthoryear{Bowman \bgroup et al\mbox.\egroup
  }{2016}]{bowman2016generating}
Bowman, S.~R.; Vilnis, L.; Vinyals, O.; Dai, A.~M.; Jozefowicz, R.; and Bengio,
  S.
\newblock 2016.
\newblock Generating sentences from a continuous space.
\newblock {\em CoNLL 2016} ~10.

\bibitem[\protect\citeauthoryear{Chen \bgroup et al\mbox.\egroup
  }{2018}]{chen2018variational}
Chen, W.; Xiong, W.; Yan, X.; and Wang, W.~Y.
\newblock 2018.
\newblock Variational knowledge graph reasoning.
\newblock In {\em Proceedings of the 2018 Conference of the North American
  Chapter of the Association for Computational Linguistics: Human Language
  Technologies, Volume 1 (Long Papers)}, volume~1,  1823--1832.

\bibitem[\protect\citeauthoryear{Dai \bgroup et al\mbox.\egroup
  }{2018}]{dai2018syntax}
Dai, H.; Tian, Y.; Dai, B.; Skiena, S.; and Song, L.
\newblock 2018.
\newblock Syntax-directed variational autoencoder for structured data.
\newblock {\em arXiv preprint arXiv:1802.08786}.

\bibitem[\protect\citeauthoryear{Fu \bgroup et al\mbox.\egroup
  }{2018}]{fu2018style}
Fu, Z.; Tan, X.; Peng, N.; Zhao, D.; and Yan, R.
\newblock 2018.
\newblock Style transfer in text: Exploration and evaluation.
\newblock In {\em Thirty-Second AAAI Conference on Artificial Intelligence}.

\bibitem[\protect\citeauthoryear{Goodfellow \bgroup et al\mbox.\egroup
  }{2014}]{goodfellow2014generative}
Goodfellow, I.; Pouget-Abadie, J.; Mirza, M.; Xu, B.; Warde-Farley, D.; Ozair,
  S.; Courville, A.; and Bengio, Y.
\newblock 2014.
\newblock Generative adversarial nets.
\newblock In {\em Advances in neural information processing systems},
  2672--2680.

\bibitem[\protect\citeauthoryear{Gregor \bgroup et al\mbox.\egroup
  }{2015}]{gregor2015draw}
Gregor, K.; Danihelka, I.; Graves, A.; Rezende, D.; and Wierstra, D.
\newblock 2015.
\newblock Draw: A recurrent neural network for image generation.
\newblock In {\em International Conference on Machine Learning},  1462--1471.

\bibitem[\protect\citeauthoryear{Hochreiter and
  Schmidhuber}{1997}]{hochreiter1997long}
Hochreiter, S., and Schmidhuber, J.
\newblock 1997.
\newblock Long short-term memory.
\newblock {\em Neural computation} 9(8):1735--1780.

\bibitem[\protect\citeauthoryear{Hu \bgroup et al\mbox.\egroup
  }{2017}]{hu2017toward}
Hu, Z.; Yang, Z.; Liang, X.; Salakhutdinov, R.; and Xing, E.~P.
\newblock 2017.
\newblock Toward controlled generation of text.
\newblock In {\em International Conference on Machine Learning},  1587--1596.

\bibitem[\protect\citeauthoryear{Hyyr{\"o}}{2001}]{hyyro2001explaining}
Hyyr{\"o}, H.
\newblock 2001.
\newblock Explaining and extending the bit-parallel approximate string matching
  algorithm of myers.
\newblock Technical report, Citeseer.

\bibitem[\protect\citeauthoryear{Iyyer \bgroup et al\mbox.\egroup
  }{2018}]{iyyer2018adversarial}
Iyyer, M.; Wieting, J.; Gimpel, K.; and Zettlemoyer, L.
\newblock 2018.
\newblock Adversarial example generation with syntactically controlled
  paraphrase networks.
\newblock In {\em Proceedings of the 2018 Conference of the North American
  Chapter of the Association for Computational Linguistics: Human Language
  Technologies, Volume 1 (Long Papers)}, volume~1,  1875--1885.

\bibitem[\protect\citeauthoryear{Kim \bgroup et al\mbox.\egroup
  }{2018}]{kim2018semi}
Kim, Y.; Wiseman, S.; Miller, A.; Sontag, D.; and Rush, A.
\newblock 2018.
\newblock Semi-amortized variational autoencoders.
\newblock In {\em International Conference on Machine Learning},  2683--2692.

\bibitem[\protect\citeauthoryear{Kingma and Ba}{2014}]{kingma2014adam}
Kingma, D.~P., and Ba, J.
\newblock 2014.
\newblock Adam: A method for stochastic optimization.
\newblock In {\em International Conference on Learning Representations (ICLR)}.

\bibitem[\protect\citeauthoryear{Kingma and Welling}{2013}]{kingma2013auto}
Kingma, D.~P., and Welling, M.
\newblock 2013.
\newblock Auto-encoding variational bayes.
\newblock In {\em Proceedings of the 2nd International Conference on Learning
  Representations (ICLR)}.

\bibitem[\protect\citeauthoryear{Kingma \bgroup et al\mbox.\egroup
  }{2014}]{kingma2014semi}
Kingma, D.~P.; Mohamed, S.; Rezende, D.~J.; and Welling, M.
\newblock 2014.
\newblock Semi-supervised learning with deep generative models.
\newblock In {\em Advances in Neural Information Processing Systems},
  3581--3589.

\bibitem[\protect\citeauthoryear{Kusner, Paige, and
  Hern{\'a}ndez-Lobato}{2017}]{kusner2017grammar}
Kusner, M.~J.; Paige, B.; and Hern{\'a}ndez-Lobato, J.~M.
\newblock 2017.
\newblock Grammar variational autoencoder.
\newblock In {\em Proceedings of the 34th International Conference on Machine
  Learning-Volume 70},  1945--1954.
\newblock JMLR. org.

\bibitem[\protect\citeauthoryear{Li \bgroup et al\mbox.\egroup
  }{2018}]{li2018unified}
Li, Z.; He, S.; Cai, J.; Zhang, Z.; Zhao, H.; Liu, G.; Li, L.; and Si, L.
\newblock 2018.
\newblock A unified syntax-aware framework for semantic role labeling.
\newblock In {\em Proceedings of the 2018 Conference on Empirical Methods in
  Natural Language Processing},  2401--2411.

\bibitem[\protect\citeauthoryear{Liu \bgroup et al\mbox.\egroup
  }{2018}]{liu2018treegan}
Liu, X.; Kong, X.; Liu, L.; and Chiang, K.
\newblock 2018.
\newblock Treegan: Syntax-aware sequence generation with generative adversarial
  networks.
\newblock In {\em 2018 IEEE International Conference on Data Mining (ICDM)},
  1140--1145.
\newblock IEEE.

\bibitem[\protect\citeauthoryear{Maaten and
  Hinton}{2008}]{maaten2008visualizing}
Maaten, L. v.~d., and Hinton, G.
\newblock 2008.
\newblock Visualizing data using t-sne.
\newblock {\em Journal of machine learning research} 9(Nov):2579--2605.

\bibitem[\protect\citeauthoryear{Mansimov \bgroup et al\mbox.\egroup
  }{2015}]{mansimov2015generating}
Mansimov, E.; Parisotto, E.; Ba, J.~L.; and Salakhutdinov, R.
\newblock 2015.
\newblock Generating images from captions with attention.
\newblock In {\em International Conference on Learning Representations (ICLR)}.

\bibitem[\protect\citeauthoryear{Marcus, Marcinkiewicz, and
  Santorini}{1993}]{marcus1993building}
Marcus, M.~P.; Marcinkiewicz, M.~A.; and Santorini, B.
\newblock 1993.
\newblock Building a large annotated corpus of english: The penn treebank.
\newblock {\em Computational linguistics} 19(2):313--330.

\bibitem[\protect\citeauthoryear{Miao, Yu, and Blunsom}{2016}]{miao2016neural}
Miao, Y.; Yu, L.; and Blunsom, P.
\newblock 2016.
\newblock Neural variational inference for text processing.
\newblock In {\em International Conference on Machine Learning},  1727--1736.

\bibitem[\protect\citeauthoryear{Mikolov \bgroup et al\mbox.\egroup
  }{2011}]{mikolov2011extensions}
Mikolov, T.; Kombrink, S.; Burget, L.; {\v{C}}ernock{\`y}, J.; and Khudanpur,
  S.
\newblock 2011.
\newblock Extensions of recurrent neural network language model.
\newblock In {\em 2011 IEEE International Conference on Acoustics, Speech and
  Signal Processing (ICASSP)},  5528--5531.
\newblock IEEE.

\bibitem[\protect\citeauthoryear{Mueller, Gifford, and
  Jaakkola}{2017}]{mueller2017sequence}
Mueller, J.; Gifford, D.; and Jaakkola, T.
\newblock 2017.
\newblock Sequence to better sequence: continuous revision of combinatorial
  structures.
\newblock In {\em Proceedings of the 34th International Conference on Machine
  Learning-Volume 70},  2536--2544.
\newblock JMLR. org.

\bibitem[\protect\citeauthoryear{Rabinovich, Stern, and
  Klein}{2017}]{rabinovich2017abstract}
Rabinovich, M.; Stern, M.; and Klein, D.
\newblock 2017.
\newblock Abstract syntax networks for code generation and semantic parsing.
\newblock In {\em Proceedings of the 55th Annual Meeting of the Association for
  Computational Linguistics (Volume 1: Long Papers)},  1139--1149.

\bibitem[\protect\citeauthoryear{Radford, Metz, and
  Chintala}{2015}]{radford2015unsupervised}
Radford, A.; Metz, L.; and Chintala, S.
\newblock 2015.
\newblock Unsupervised representation learning with deep convolutional
  generative adversarial networks.
\newblock {\em arXiv preprint arXiv:1511.06434}.

\bibitem[\protect\citeauthoryear{Rezende and
  Mohamed}{2015}]{rezende2015variational}
Rezende, D., and Mohamed, S.
\newblock 2015.
\newblock Variational inference with normalizing flows.
\newblock In {\em International Conference on Machine Learning},  1530--1538.

\bibitem[\protect\citeauthoryear{Robbins and
  Monro}{1985}]{robbins1985stochastic}
Robbins, H., and Monro, S.
\newblock 1985.
\newblock A stochastic approximation method.
\newblock In {\em Herbert Robbins Selected Papers}. Springer.
\newblock  102--109.

\bibitem[\protect\citeauthoryear{Serban \bgroup et al\mbox.\egroup
  }{2017}]{serban2017hierarchical}
Serban, I.~V.; Sordoni, A.; Lowe, R.; Charlin, L.; Pineau, J.; Courville,
  A.~C.; and Bengio, Y.
\newblock 2017.
\newblock A hierarchical latent variable encoder-decoder model for generating
  dialogues.
\newblock In {\em AAAI},  3295--3301.

\bibitem[\protect\citeauthoryear{Shen \bgroup et al\mbox.\egroup
  }{2017}]{shen2017style}
Shen, T.; Lei, T.; Barzilay, R.; and Jaakkola, T.
\newblock 2017.
\newblock Style transfer from non-parallel text by cross-alignment.
\newblock In {\em Advances in neural information processing systems},
  6830--6841.

\bibitem[\protect\citeauthoryear{Sohn, Lee, and Yan}{2015}]{sohn2015learning}
Sohn, K.; Lee, H.; and Yan, X.
\newblock 2015.
\newblock Learning structured output representation using deep conditional
  generative models.
\newblock In {\em Advances in Neural Information Processing Systems},
  3483--3491.

\bibitem[\protect\citeauthoryear{Srivastava and
  Sutton}{2017}]{srivastava2017autoencoding}
Srivastava, A., and Sutton, C.
\newblock 2017.
\newblock Autoencoding variational inference for topic models.
\newblock In {\em International Conference on Learning Representations (ICLR)}.

\bibitem[\protect\citeauthoryear{Strubell \bgroup et al\mbox.\egroup
  }{2018}]{strubell2018linguistically}
Strubell, E.; Verga, P.; Andor, D.; Weiss, D.; and McCallum, A.
\newblock 2018.
\newblock Linguistically-informed self-attention for semantic role labeling.
\newblock In {\em Proceedings of the 2018 Conference on Empirical Methods in
  Natural Language Processing},  5027--5038.

\bibitem[\protect\citeauthoryear{Xiao, Zhao, and
  Wang}{2018}]{xiao2018dirichlet}
Xiao, Y.; Zhao, T.; and Wang, W.~Y.
\newblock 2018.
\newblock Dirichlet variational autoencoder for text modeling.
\newblock {\em arXiv preprint arXiv:1811.00135}.

\bibitem[\protect\citeauthoryear{Yan \bgroup et al\mbox.\egroup
  }{2016}]{yan2016attribute2image}
Yan, X.; Yang, J.; Sohn, K.; and Lee, H.
\newblock 2016.
\newblock Attribute2image: Conditional image generation from visual attributes.
\newblock In {\em European Conference on Computer Vision},  776--791.
\newblock Springer.

\bibitem[\protect\citeauthoryear{Yang \bgroup et al\mbox.\egroup
  }{2017}]{yang2017improved}
Yang, Z.; Hu, Z.; Salakhutdinov, R.; and Berg-Kirkpatrick, T.
\newblock 2017.
\newblock Improved variational autoencoders for text modeling using dilated
  convolutions.
\newblock In {\em International Conference on Machine Learning},  3881--3890.

\bibitem[\protect\citeauthoryear{Yang \bgroup et al\mbox.\egroup
  }{2018}]{yang2018unsupervised}
Yang, Z.; Hu, Z.; Dyer, C.; Xing, E.~P.; and Berg-Kirkpatrick, T.
\newblock 2018.
\newblock Unsupervised text style transfer using language models as
  discriminators.
\newblock In {\em Advances in Neural Information Processing Systems},
  7298--7309.

\bibitem[\protect\citeauthoryear{Yin and Neubig}{2017}]{yin2017syntactic}
Yin, P., and Neubig, G.
\newblock 2017.
\newblock A syntactic neural model for general-purpose code generation.
\newblock In {\em Proceedings of the 55th Annual Meeting of the Association for
  Computational Linguistics (Volume 1: Long Papers)},  440--450.

\bibitem[\protect\citeauthoryear{Yu and Koltun}{2016}]{yu2016multi}
Yu, F., and Koltun, V.
\newblock 2016.
\newblock Multi-scale context aggregation by dilated convolutions.
\newblock In {\em International Conference on Learning Representations (ICLR)}.

\bibitem[\protect\citeauthoryear{Zhang \bgroup et al\mbox.\egroup
  }{2016}]{zhang2016variational}
Zhang, B.; Xiong, D.; Duan, H.; Zhang, M.; et~al.
\newblock 2016.
\newblock Variational neural machine translation.
\newblock In {\em Proceedings of the 2016 Conference on Empirical Methods in
  Natural Language Processing},  521--530.

\bibitem[\protect\citeauthoryear{Zhang \bgroup et al\mbox.\egroup
  }{2018}]{zhang2018variational}
Zhang, Y.; Dai, H.; Kozareva, Z.; Smola, A.~J.; and Song, L.
\newblock 2018.
\newblock Variational reasoning for question answering with knowledge graph.
\newblock In {\em Thirty-Second AAAI Conference on Artificial Intelligence}.

\bibitem[\protect\citeauthoryear{Zhao \bgroup et al\mbox.\egroup
  }{2018}]{zhao2018adversarially}
Zhao, J.; Kim, Y.; Zhang, K.; Rush, A.~M.; and LeCun, Y.
\newblock 2018.
\newblock Adversarially regularized autoencoders.
\newblock In {\em 35th International Conference on Machine Learning, ICML
  2018},  9405--9420.
\newblock International Machine Learning Society (IMLS).

\bibitem[\protect\citeauthoryear{Zhao, Zhao, and
  Eskenazi}{2017}]{zhao2017learning}
Zhao, T.; Zhao, R.; and Eskenazi, M.
\newblock 2017.
\newblock Learning discourse-level diversity for neural dialog models using
  conditional variational autoencoders.
\newblock In {\em Proceedings of the 55th Annual Meeting of the Association for
  Computational Linguistics (Volume 1: Long Papers)}, volume~1,  654--664.

\bibitem[\protect\citeauthoryear{Zhu \bgroup et al\mbox.\egroup
  }{2015}]{zhu2015aligning}
Zhu, Y.; Kiros, R.; Zemel, R.; Salakhutdinov, R.; Urtasun, R.; Torralba, A.;
  and Fidler, S.
\newblock 2015.
\newblock Aligning books and movies: Towards story-like visual explanations by
  watching movies and reading books.
\newblock In {\em Proceedings of the IEEE international conference on computer
  vision},  19--27.

\end{thebibliography}

\end{document}